\documentclass[10pt,twocolumn,letterpaper]{article}

\usepackage[pagenumbers]{cvpr} 

\usepackage{colortbl}
\usepackage{multirow}
\usepackage{placeins}
\usepackage{pifont}
\usepackage{tcolorbox}
\tcbuselibrary{listings}
\usepackage{bibunits}

\definecolor{cvprblue}{rgb}{0.21,0.49,0.74}
\usepackage[pagebackref,breaklinks,colorlinks,allcolors=cvprblue]{hyperref}

\def\confName{CVPR}
\def\confYear{2027}

\title{InstructMontage: Beat-Aware Learning for Instruction-Guided \\ Multi-Track Video Editing}

\author{%
Zhengtong Zhu$^{1,2}$ \quad Wenjie Dong$^{1}$ \quad Zhi Xu$^{1}$ \quad He Wang$^{1}$ \quad Yudong Zhu$^{1,\dagger}$\\
Xin Ouyang$^{1}$ \quad Chen Zhu$^{1}$ \quad Jiaqing Fan$^{2,\ddagger}$ \quad Liwei Qian$^{1}$ \quad Xin Pei$^{1}$\\[0.4em]
$^{1}$Baidu (baidu.com) \quad $^{2}$Soochow University (suda.edu.cn)\\
{\tt\small zhuzhengtong@baidu.com \quad jqfan@suda.edu.cn}%
}

\begin{document}
\twocolumn[{%
    \renewcommand\twocolumn[1][]{##1}%
    \maketitle
    \centering
    \includegraphics[width=0.95\textwidth]{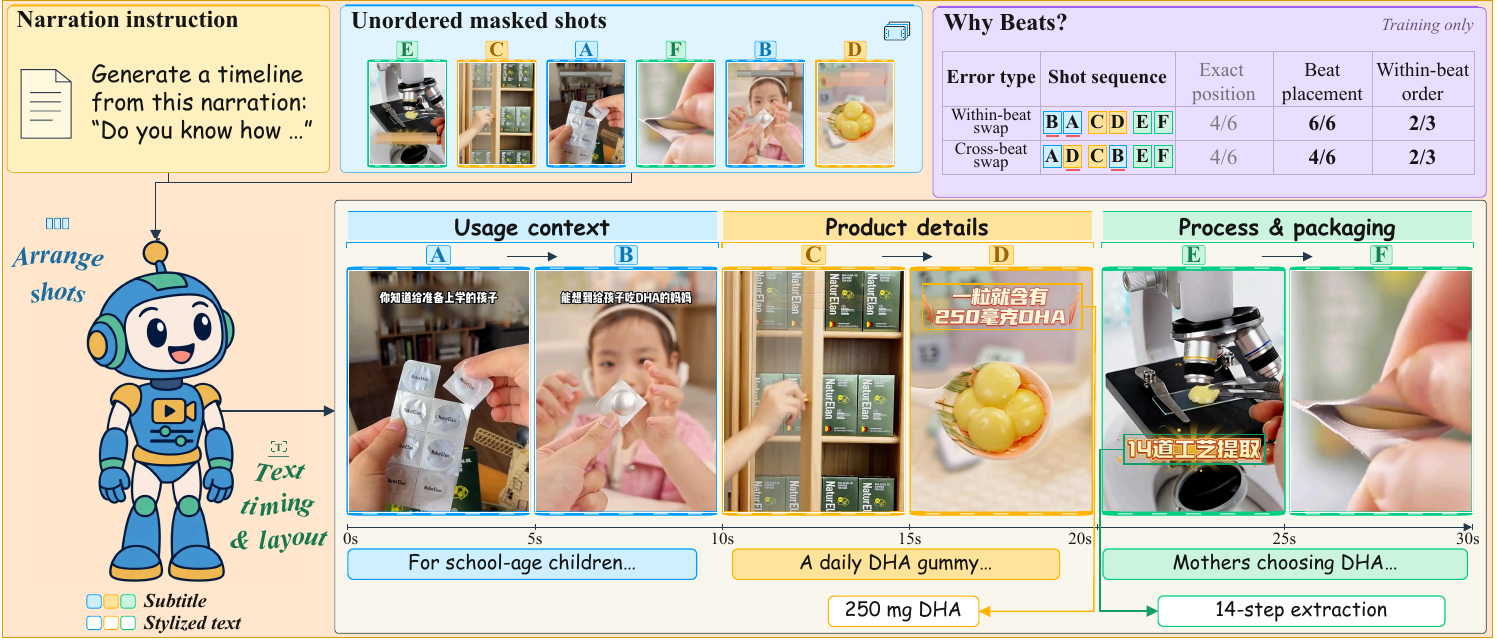}
    \captionof{figure}{\textbf{Overview of instruction-guided multi-track video editing.} InstructMontage maps narration and unordered shots to video, subtitle, and stylized-text tracks. Training-only beat supervision distinguishes ordering errors with identical exact-position scores.}
    \label{fig:intro}
    \vspace{1em}
}]
\iftoggle{cvprfinal}{%
  \begingroup
    \renewcommand{\thefootnote}{\fnsymbol{footnote}}
    \footnotetext[2]{Project leader.}
    \footnotetext[3]{Corresponding author.}
  \endgroup
}{}
\begin{abstract}
Instruction-guided video editing requires organizing visual content into a coherent narrative while coordinating multiple tracks on a shared timeline.
Learning this capability is challenging: exact-position rewards provide limited guidance for shot ordering, processing numerous shots incurs visual token costs, and existing video annotations offer limited supervision for structured multi-track editing.
We introduce \textbf{InstructMontage}, a unified vision-language model to jointly generate shot arrangements, subtitles, and stylized text within a single executable multi-track timeline from editing instructions and source shots.
At its core is \textbf{Beat-Aware Hierarchical Reward (BAHR)}, which uses narrative beats as an intermediate level of supervision between individual shots and the complete sequence.
Derived from reference edits, pseudo beats support complementary rewards for beat-level grouping and within-beat relative ordering, alongside an exact-position objective.
These signals guide narrative organization without requiring beat annotations at inference time.
To reduce visual input costs, we develop \textbf{Instruction-Aware Shot Compression (IASC)}, which uses a lightweight token selector distilled from a frozen model's instruction-to-visual attention to retain native visual tokens and their spatiotemporal positions.
We construct \textbf{MTVE-Bench} (Multi-Track Video Editing Benchmark), featuring aligned video, subtitle, and stylized-text annotations.
Through supervised fine-tuning followed by group-relative policy optimization, InstructMontage learns to produce structured timelines that can be deterministically rendered into edited videos.
We connect narrative-level learning, efficient multi-shot understanding, and executable multi-track editing within a unified framework.
\end{abstract}
    
\section{Introduction}
\label{sec:intro}

Video editing turns a collection of shots into a purposeful narrative.
Following an instruction requires more than retrieving footage: an editor must arrange shots, align subtitles, and place emphatic text on a shared timeline.
These decisions span temporal and spatial granularities.

Automated editing has progressed from shot assembly to language-conditioned timeline generation.
RATV~\cite{ratv} models script-conditioned retrieval and assembly, TGT~\cite{tgt} learns movie-to-trailer sequencing, and Timeline Assembler~\cite{timelineassembler} modifies visual sequences according to instructions.
These formulations organize footage, but their shot-sequence outputs do not encompass the object-level decisions needed to coordinate video and overlaid text.
Recent generative systems extend beyond shot ordering.
AutoCut~\cite{autocut} unifies video selection, ordering, script generation, and music selection, with subtitles derived from script sentences during rendering.
We focus on directly predicting video, subtitle, and stylized-text objects.
Text content, display intervals, and stylized-text placement become explicit supervision and evaluation targets, rather than decisions left entirely to downstream composition.

A complementary line automates editing through agents and workflows.
LAVE~\cite{lave} supports language-based planning and editing actions, EditDuet~\cite{editduet} refines timelines through Editor--Critic interaction, and VideoAgent~\cite{videoagent} orchestrates specialized agents.
Our focus is instead on learning a direct mapping from instructions and shots to a unified timeline within a single vision-language model, rather than organizing editing-tool interactions.

Three challenges remain.
First, heterogeneous timeline objects require aligned supervision for video arrangement and text timing and placement.
AVE~\cite{ave} and VEU-Bench~\cite{veu-bench} provide valuable cinematographic annotations and editing-understanding tasks, but are not designed to supervise this joint output.
Second, exact shot positions do not fully express narrative organization.
Position-based scores provide partial credit, yet can assign identical rewards to arrangements with different local structures.
Third, numerous source shots introduce substantial visual token costs, motivating compression that preserves instruction-relevant evidence and temporal structure.

We introduce \textbf{InstructMontage}, a unified framework for predicting executable timelines with video, subtitle, and stylized-text tracks.
We establish \textbf{MTVE-Bench} (Multi-Track Video Editing Benchmark), providing aligned multi-track annotations for supervised learning and evaluation.

Motivated by redundancy reduction and query-aware compression in LongVU~\cite{longvu} and PruneVid~\cite{prunevid}, we develop \textbf{Instruction-Aware Shot Compression (IASC)} to reduce visual input costs.
For multi-shot editing, IASC distills instruction-to-visual attention into a selector that retains native visual tokens and their spatiotemporal positions.

Our central contribution is \textbf{Beat-Aware Hierarchical Reward (BAHR)}.
Narrative beats provide an intermediate level between individual shots and the complete sequence, helping distinguish coarse narrative organization from fine-grained ordering.
For example, placing related shots in the correct narrative phase can represent useful progress even when their exact positions remain incorrect.
Reference-derived pseudo beats support complementary rewards for beat-level grouping and within-beat ordering alongside exact-position supervision.
We use these signals to guide group-relative policy optimization~\cite{grpo}.
The rewards supervise video-shot organization and require no beat annotations at inference.

We summarize our contributions as follows:
\begin{itemize}
    \item We introduce \textbf{InstructMontage}, a unified framework that maps editing instructions and source shots to executable timelines with video, subtitle, and stylized-text tracks.
    \item Since scaling to many source shots incurs substantial visual-token costs, we develop \textbf{IASC}, an instruction-conditioned selector that retains native visual tokens.
    \item Exact-position objectives alone inadequately capture narrative structure; we therefore propose \textbf{BAHR}, combining supervision for beat-level grouping, within-beat ordering, and exact positions.
    \item To address the lack of multi-track supervision, we construct \textbf{MTVE-Bench} with aligned video and text annotations, and evaluate InstructMontage on both MTVE-Bench and the public VSPD dataset~\cite{ratv}.
\end{itemize}

\begin{figure*}[t]
    \centering
    \includegraphics[width=0.85\textwidth]{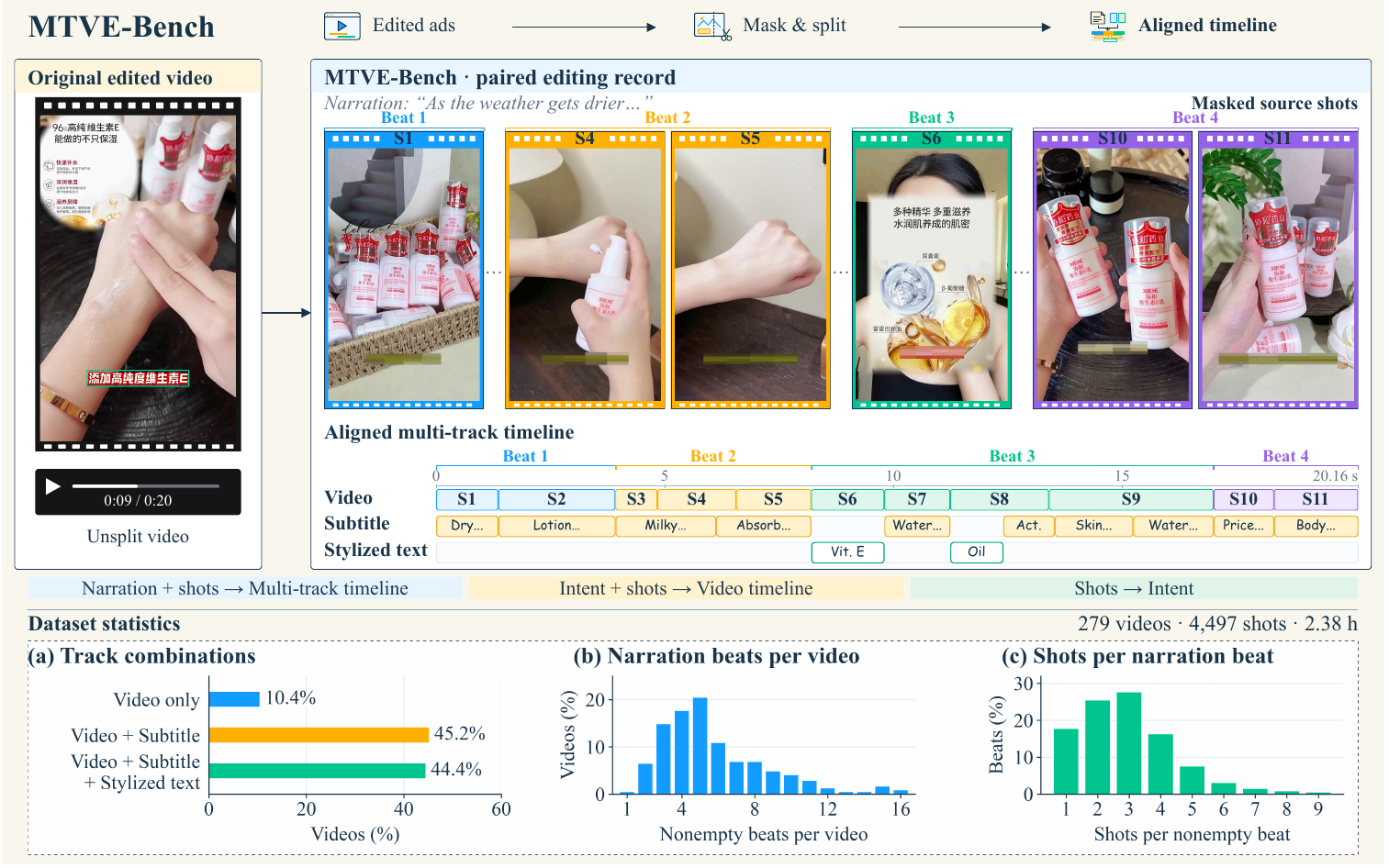}
    \caption{\textbf{Overview of MTVE-Bench.} Edited advertisements are decomposed into masked source shots and aligned video, subtitle, and stylized-text annotations. The schematic example illustrates shot organization and text timing and placement, with colors indicating reference-derived semantic beats. Three tasks cover narration-guided multi-track editing, intent-guided ordering, and intent understanding. Panels (a--c) summarize track combinations, narration beats per video, and shots per narration beat in the current curated snapshot.}
    \label{fig:benchmark}
\end{figure*}

\section{Related Work}
\label{sec:related_work}

\subsection{Video Generation and Editing}
FilmWeaver~\cite{filmweaver} generates consistent multi-shot videos using cache-guided autoregressive diffusion, whereas Uni-ViGU~\cite{univigu} unifies video generation and understanding through a diffusion-based generator.
Xue et al.~\cite{human-motion-survey} survey human-motion synthesis across visual, textual, and audio conditioning; related capabilities include portrait animation~\cite{liveportrait}, video dubbing~\cite{musetalk}, streaming avatars~\cite{soulx-flashtalk,soulx-flashhead}, multi-person interaction~\cite{anytalker}, and face swapping~\cite{canonswap}.
These capabilities synthesize or modify subjects within footage, complementing editing from existing assets.
Write-A-Video~\cite{write-a-video} and RATV~\cite{ratv} study text-guided retrieval and assembly, while SKALD~\cite{skald} learns sequence coherence for beam-search shot assembly.
Timeline Assembler~\cite{timelineassembler} performs instruction-conditioned sequence modification.
AutoCut~\cite{autocut} jointly models video, scripts, and music, deriving subtitles from script sentences during rendering.
LAVE~\cite{lave}, EditDuet~\cite{editduet}, and VideoAgent~\cite{videoagent} support editing through interactive assistance, editing--critique feedback, and specialized-agent orchestration, respectively.
InstructMontage directly predicts unified timelines of video, subtitle, and stylized-text objects, making text timing and placement explicit learning targets rather than relying on script-derived overlays or tool workflows.

\subsection{Video Editing Benchmarks}
AVE~\cite{ave} supplies cinematographic annotations; VSPD~\cite{ratv} evaluates script-conditioned retrieval and ordering; VEU-Bench~\cite{veu-bench} assesses recognition, reasoning, and judgment of editing attributes; and MSV3C~\cite{skald} provides multi-shot data with human ratings of coherence and semantic consistency.
DNTextSpotter~\cite{dntextspotter} addresses arbitrary-shaped scene-text localization and recognition, rather than supervision of editor-defined text objects.
MTVE-Bench complements these resources with aligned video, subtitle, and stylized-text annotations, evaluating shot arrangement together with text content, visibility intervals, and spatial placement on a shared timeline.

\subsection{Efficient Multimodal Representations}
FastV~\cite{fastv} uses early-layer attention for pruning, while VisionZip~\cite{visionzip} selects informative visual tokens; LongVU~\cite{longvu} and PruneVid~\cite{prunevid} reduce spatiotemporal redundancy in video.
AIM~\cite{aim} aggregates multimodal demonstrations into latent representations associated with textual labels, reducing retained demonstration visual tokens.
Forcing-KV~\cite{forcing-kv} combines static and dynamic pruning of historical KV caches in autoregressive video diffusion, whereas MJev-Spatial~\cite{mjev-spatial} scores spatial candidates using pretrained query/key projections without autoregressively generating answer tokens.
These approaches target input redundancy, demonstration costs, generation memory, and decision decoding, respectively.
IASC selects instruction-relevant native visual tokens, retaining vectors, shot boundaries, and spatiotemporal positions.

\subsection{Group-Relative Policy Optimization (GRPO)}
GRPO~\cite{grpo} replaces PPO's~\cite{ppo} learned critic with group-relative feedback from responses sampled for the same prompt.
Dr.~GRPO~\cite{dr-grpo} revises normalization to reduce length and difficulty biases, while DAPO~\cite{dapo} improves clipping, sampling, and loss aggregation.
KVPO~\cite{kvpo} extends group-relative optimization to ODE-native video generation through KV semantic exploration.
Unlike step-level reasoning supervision~\cite{prm800k}, BAHR evaluates beat grouping and within-beat order in complete editing plans.

\section{MTVE-Bench}
\label{sec:benchmark}

\subsection{Task Definition}
\label{sec:benchmark-task}

Given an instruction $I$ and unordered source shots $\mathcal S=\{s_k\}_{k=1}^{N}$, the target is a shared timeline
\begin{equation}
    \mathcal T=\big(\mathcal T_{\mathrm{vid}},
    \mathcal T_{\mathrm{sub}},\mathcal T_{\mathrm{sty}}\big).
    \label{eq:benchmark-timeline}
\end{equation}
Video objects specify asset identifiers and source/timeline intervals; subtitle objects contain text and display intervals; stylized-text objects additionally carry normalized spatial boxes.
Intervals are half-open and expressed in integer milliseconds.
Each source shot appears once, at full length, so the task concerns ordering and text-object composition.

The primary task maps narration and shuffled shots to a multi-track timeline.
Two auxiliary tasks map marketing intent and shuffled shots to a video-only timeline, and reference-ordered shots to marketing intent, respectively.
They isolate intent-conditioned ordering and intent understanding from joint multi-track editing.
Further dataset details are provided in Supplementary Sec.~\ref{sec:supp-benchmark}.

\subsection{Dataset Statistics and Quality Control}
\label{sec:benchmark-statistics}

The current snapshot, curated from E-VAds~\cite{evads}, contains 279 videos, 4,497 shots, and 2.38 hours of footage.
Video-only, video--subtitle, and video--subtitle--stylized-text annotations account for 10.4\%, 45.2\%, and 44.4\% of videos, respectively.
Narration annotations cover 250 videos and 1,432 nonempty narration beats; Fig.~\ref{fig:benchmark}(b,c) reports their structural distributions.
Automatic checks verify temporal bounds, shot coverage, nonempty text, and valid boxes.
Manual review supports correcting text categories, content, timing, geometry, and semantic annotations.
We split the 279 source videos into 251 training videos and 28 test videos, keeping all derived shots and task instances from each source video in the same split.

\subsection{Evaluation Protocol}
\label{sec:benchmark-evaluation}

For predicted shot IDs $\hat p_j$ and reference IDs $p_j^\star$, Shot Matching Score (SMS)~\cite{ratv} measures exact-position agreement:
\begin{equation}
    \mathrm{SMS}=\frac1N\sum_{j=1}^{N}
    \mathbf 1[\hat p_j=p_j^\star].
    \label{eq:benchmark-sms}
\end{equation}
Missing positions contribute zero; the denominator remains the reference length $N$.

Let $w(u)$ denote the whitespace-stripped text of object $u$.
Character similarity is
\begin{equation}
    c(u,v)=1-\frac{d_{\mathrm{edit}}(w(u),w(v))}
    {\max\{|w(u)|,|w(v)|\}},
    \label{eq:benchmark-text-similarity}
\end{equation}
where $d_{\mathrm{edit}}$ is Levenshtein distance, with $c=1$ for two empty strings.
Pairwise qualities combine text with temporal and spatial alignment:
\begin{equation}
    \begin{aligned}
        q_{\mathrm{sub}}(u,v)&=c(u,v)\operatorname{tIoU}(u,v),\\[4pt]
        q_{\mathrm{sty}}(u,v)&=q_{\mathrm{sub}}(u,v)
        \operatorname{bIoU}(u,v),
    \end{aligned}
    \label{eq:benchmark-text-quality}
\end{equation}
where tIoU and bIoU are interval and box intersection-over-union; invalid intervals or boxes receive zero corresponding IoU.
Descending-quality greedy one-to-one matching yields $\mathcal M_k$ for text track $k\in\{\mathrm{sub},\mathrm{sty}\}$.
Ties are resolved by prediction and reference indices.
Its Soft-F1 is
\begin{equation}
    F_k=\frac{2\sum_{(u,v)\in\mathcal M_k}q_k(u,v)}
    {|\widehat{\mathcal T}_k|+|\mathcal T_k^\star|}.
    \label{eq:benchmark-soft-f1}
\end{equation}
Two empty tracks score one; exactly one empty track scores zero.
The primary multi-track score is
\begin{equation}
    S_{\mathrm{MT}}=
    \frac{\mathrm{SMS}+F_{\mathrm{sub}}+F_{\mathrm{sty}}}{3}.
    \label{eq:benchmark-multitrack-score}
\end{equation}
Intent-guided ordering uses SMS.
For whitespace-stripped predicted and reference intent strings $\hat z,z^\star$, character-level ROUGE-L F1~\cite{rouge} is
\begin{equation}
    F_{\mathrm{RL}}=
    \frac{2\operatorname{LCS}(\hat z,z^\star)}{|\hat z|+|z^\star|},
    \label{eq:benchmark-rouge}
\end{equation}
where LCS denotes the longest common subsequence length.
We macro-average each task's primary score over its samples, then equally average enabled task means.
JSON parsing and timeline validity are reported separately.
Preparation, generation, or parsing failures remain in the denominator with zero task scores.
Parsed components are scored independently of full-timeline validity and generation stop reason.
These scores are distinct from the RL reward.

\section{Method}
\label{sec:method}

\begin{figure*}[t]
    \centering
    \includegraphics[width=\textwidth]{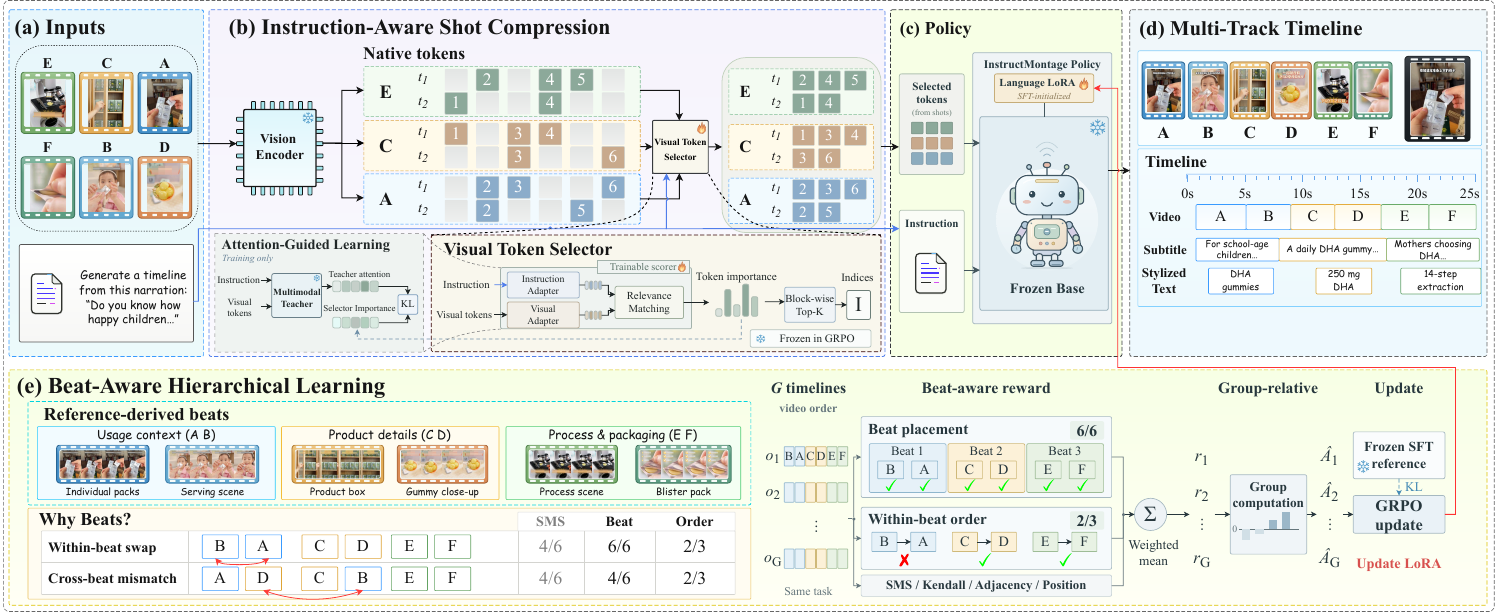}
    \caption{\textbf{Overview of InstructMontage.} (a) An editing instruction conditions shuffled source shots. (b) IASC selects native visual tokens within temporal blocks, preserving their vectors and spatiotemporal positions. A frozen attention teacher distills instruction-to-visual relevance into the lightweight selector during SFT; the selector is frozen during GRPO. (c) The editing policy combines a frozen base with SFT-initialized language LoRA. (d) It generates a unified timeline with video, subtitle, and stylized-text tracks. (e) BAHR evaluates beat placement and within-beat order for $G$ sampled complete plans, alongside basic ordering signals. Group-relative advantages and KL regularization against a frozen SFT reference update only language LoRA. The attention teacher and policy reference serve distinct roles.}
    \label{fig:pipeline}
\end{figure*}

\subsection{Framework Overview}
\label{sec:overview}

As illustrated in Fig.~\ref{fig:pipeline}, InstructMontage maps an editing instruction $I$ and an unordered set of source shots $\mathcal S=\{s_k\}_{k=1}^{N}$ to an executable multi-track timeline.
Each shot has a unique asset identifier and retains its own temporal boundaries.
A frozen visual encoder and Merger, denoted jointly by $E_{\mathrm{vis}}$, extract native visual tokens; IASC selects instruction-relevant tokens using parameters $\phi$:
\begin{equation}
    V_k=E_{\mathrm{vis}}(s_k),\qquad
    \widetilde V_k=C_\phi(V_k,I).
    \label{eq:visual-input}
\end{equation}
The selected tokens retain their original vectors and spatiotemporal positions.
We denote by $x$ the resulting model input, comprising the instruction, selected visual tokens, asset identifiers, and temporal and positional metadata.
Shot boundaries remain explicit, allowing the model to associate visual evidence with the corresponding source asset.

The editing policy $\pi_\theta$, built on Qwen3.5, generates a serialized timeline $y=(y_1,\ldots,y_L)$ autoregressively:
\begin{equation}
    \pi_\theta(y\mid x)
    =\prod_{t=1}^{L}\pi_\theta(y_t\mid x,y_{<t}),
    \label{eq:timeline-policy}
\end{equation}
where $\theta$ denotes the language LoRA~\cite{lora} parameters and the pretrained base weights are fixed.
Parsing $y$ yields a timeline $\mathcal T$ with video, subtitle, and stylized-text tracks.
Video arrangement, text content, display intervals, and stylized-text placement are represented on a shared temporal axis.

Training begins with Supervised Fine-Tuning (SFT) on reference timelines to establish structured editing capability.
In this stage, attention distillation from a frozen teacher trains the IASC selector, while timeline supervision updates the language LoRA.
We then initialize reinforcement learning from the SFT policy, freeze the visual modules and selector, and further optimize the language LoRA.
BAHR supplies hierarchical supervision for video-shot organization.
Reference timelines and pseudo beats provide training supervision only; inference requires the editing instruction and source shots.

\subsection{GRPO}
\label{sec:grpo}

GRPO~\cite{grpo} compares complete plans for the same input without a learned critic.
A rollout policy $\pi_{\mathrm{old}}$ samples $G\geq2$ timelines $y_i$, each receiving $r_i=R(y_i;x,a^\star)$.
Reference annotations $a^\star$ enter only the reward, not the model input.
With group mean $\mu_r$ and sample standard deviation $\sigma_r$, each valid token shares the sequence advantage
\begin{equation}
    \widehat A_i=\frac{r_i-\mu_r}{\sigma_r+\delta},
    \label{eq:group-advantage}
\end{equation}
where $\delta>0$ ensures numerical stability.
We update language LoRA with the standard clipped objective and KL regularization against the frozen initial SFT reference $\pi_{\mathrm{ref}}$.
Rewards, advantages, and rollout-policy probabilities are held fixed during optimization.
The complete formulation appears in Supplementary Sec.~\ref{sec:supp-grpo}.

\subsection{Instruction-Aware Shot Compression}
\label{sec:iasc}

IASC retains instruction-relevant evidence under a per-shot visual budget.
Write $V_k=\{v_{k,b,j}\}$, where $b$ indexes $T_k$ temporal blocks and $j$ indexes spatial tokens.
Mean-pooling the frozen base's last hidden states for instruction-only input gives $z_I$.
One instruction representation is shared across its source shots, so their relevance scores are conditioned on the same editing goal.
Two trainable LayerNorm--linear projections, $f_q$ and $f_v$, map instruction and visual features to dimension $d_s$:
\begin{equation}
    e_{k,b,j}=\frac{f_q(z_I)^\top f_v(v_{k,b,j})}{\sqrt{d_s}}.
    \label{eq:iasc-relevance}
\end{equation}

To supervise relevance, a frozen base with LoRA disabled processes each shot followed by the instruction, allowing instruction queries to attend causally to visual tokens.
We extract attention from a designated full-attention layer and average heads and instruction query positions.
Let $a_{k,b,j}$ denote the resulting visual attention.
Blockwise normalization defines teacher targets, while the student uses temperature $\tau>0$:
\begin{equation}
    \begin{aligned}
        q_{k,b,j}&=\frac{a_{k,b,j}}{\sum_{\ell}a_{k,b,\ell}},\\[4pt]
        p^\phi_{k,b,j}&=\operatorname{softmax}_{j}(e_{k,b,j}/\tau).
    \end{aligned}
    \label{eq:iasc-distributions}
\end{equation}
We balance supervision across blocks, shots, and samples:
\begin{equation}
    \begin{aligned}
        \mathcal L_{\mathrm{KD}}(\phi)
        &=\mathbb E_{(I,\mathcal S)}\!\bigg[\\[4pt]
        &\frac1N\sum_{k=1}^{N}\frac1{T_k}\sum_{b=1}^{T_k}
        D_{\mathrm{KL}}(q_{k,b}\Vert p^\phi_{k,b})\bigg].
    \end{aligned}
    \label{eq:iasc-distillation}
\end{equation}
The supervised objective combines timeline cross-entropy and attention distillation:
\begin{equation}
    \mathcal L_{\mathrm{SFT}}
    =\mathcal L_{\mathrm{CE}}+\lambda_d\mathcal L_{\mathrm{KD}}.
    \label{eq:iasc-sft}
\end{equation}
Here $\lambda_d\geq0$ balances distillation.
Cross-entropy updates language LoRA $\theta$; distillation updates selector $\phi$.
Visual features, instruction features, and teacher targets are detached; hard selection does not pass cross-entropy gradients to the selector.

For shot budget $K_k$, we distribute $K_{k,b}$ nearly equally across temporal blocks, with $\sum_bK_{k,b}=K_k$ and $K_{k,b}\geq1$.
Selected indices are
\begin{equation}
    \Omega_{k,b}=\operatorname{TopK}_{j}(e_{k,b,j},K_{k,b}).
    \label{eq:iasc-selection}
\end{equation}
Budget remainders are assigned to earlier temporal blocks, and equal-score tokens follow their original spatial indices.
We gather the original token vectors in their original order, preserving shot boundaries and native three-axis M-RoPE coordinates.
Thus, compression reduces the visual context supplied to the language model without replacing native representations or bypassing visual encoding.
Inference uses frozen text encoding and the lightweight selector, without the attention teacher.
During GRPO, the selector is frozen, providing inputs for rollout and policy scoring.

\begingroup
\setlength{\aboverulesep}{0pt}
\setlength{\belowrulesep}{0pt}

\begin{table*}[tp]
    \centering
    \small
    \setlength{\tabcolsep}{10.8pt}
    \renewcommand{\arraystretch}{1.12}
    \caption{\textbf{Narration-guided multi-track editing on the MTVE-Bench test set.}
    SMS: Shot Matching Score; POS: Order Position Score; LIS: normalized longest increasing subsequence; ADJ: Order Adjacent Score; BHR: Beat Hit Rate; Sub-F1/Sty-F1: subtitle/stylized-text Soft-F1; MTS: Multi-track Score.
    Scores are in $[0,1]$ ($\uparrow$); bold indicates column-wise best scores at the displayed precision.
    RL denotes the checkpoint after one RL epoch.}
    \label{tab:narration-results}
    \begin{tabular}{l|ccccc|cc|c}
        \toprule[0.8pt]
        \rowcolor[HTML]{DBDBDB}
        & \multicolumn{5}{c|}{Shot ordering} & \multicolumn{2}{c|}{Text tracks} & Overall \\
        \rowcolor[HTML]{DBDBDB}
        \multirow{-2}{*}{Model} & SMS $\uparrow$ & POS $\uparrow$ & LIS $\uparrow$ & ADJ $\uparrow$ & BHR $\uparrow$ & Sub-F1 $\uparrow$ & Sty-F1 $\uparrow$ & MTS $\uparrow$ \\
        \midrule[0.6pt]
        Claude-opus-4.8 & 0.0944 & 0.3397 & \textbf{0.4156} & 0.1023 & 0.2545 & \textbf{0.2249} & 0.0031 & 0.1075 \\
        \rowcolor[HTML]{EEEEEE}
        Qwen3.8-27b & 0.0567 & 0.1882 & 0.2601 & 0.0884 & 0.1743 & 0.1564 & 0.0016 & 0.0715 \\
        \rowcolor[HTML]{EEEEEE}
        Kimi-k3 & 0.0517 & 0.1954 & 0.2467 & 0.0762 & 0.1408 & 0.1757 & 0.0005 & 0.0760 \\
        Gemini-3.6-flash & \textbf{0.1765} & 0.3485 & 0.3421 & \textbf{0.1399} & \textbf{0.3004} & 0.1269 & 0.0005 & 0.1013 \\
        \rowcolor[HTML]{EEEEEE}
        Gemini-3.7-flash & 0.1378 & 0.2228 & 0.2196 & 0.1130 & 0.1885 & 0.0726 & 0.0017 & 0.0707 \\
        \midrule[0.6pt]
        Qwen3.5-4B (Base) & 0.0389 & 0.1273 & 0.1710 & 0.0431 & 0.1161 & 0.0357 & 0.0009 & 0.0252 \\
        \rowcolor[HTML]{DAF7FD}
        InstructMontage (SFT) & 0.0827 & 0.3112 & 0.3276 & 0.0725 & 0.2525 & 0.1636 & 0.2692 & 0.1718 \\
        \rowcolor[HTML]{DAF7FD}
        InstructMontage (RL) & 0.0932 & \textbf{0.3844} & \textbf{0.4156} & 0.1186 & 0.2748 & 0.2236 & \textbf{0.2983} & \textbf{0.2050} \\
        \bottomrule[0.8pt]
    \end{tabular}
\end{table*}

\begin{table}[tp]
    \centering
    \small
    \setlength{\tabcolsep}{2.1pt}
    \renewcommand{\arraystretch}{1.12}
    \caption{\textbf{Intent-guided ordering on the MTVE-Bench test set.}
    Metric abbreviations and boldface follow Table~\ref{tab:narration-results}; all metrics are higher-is-better.
    RL is evaluated after one RL epoch.}
    \label{tab:intent-ordering-results}
    \begin{tabular}{l|ccccc}
        \toprule[0.8pt]
        \rowcolor[HTML]{DBDBDB}
        Model & SMS $\uparrow$ & POS $\uparrow$ & LIS $\uparrow$ & ADJ $\uparrow$ & BHR $\uparrow$ \\
        \midrule[0.6pt]
        Claude-opus-4.8 & 0.0987 & 0.3427 & 0.4148 & 0.0851 & 0.3112 \\
        \rowcolor[HTML]{EEEEEE}
        Qwen3.8-27b & 0.0788 & 0.3007 & 0.3705 & 0.0743 & 0.2828 \\
        \rowcolor[HTML]{EEEEEE}
        Kimi-k3 & 0.0823 & 0.3382 & 0.3900 & 0.0712 & 0.3189 \\
        Gemini-3.6-flash & \textbf{0.1820} & 0.3174 & 0.3313 & \textbf{0.1769} & 0.3338 \\
        \rowcolor[HTML]{EEEEEE}
        Gemini-3.7-flash & 0.1421 & 0.2283 & 0.2281 & 0.1392 & 0.2561 \\
        \midrule[0.6pt]
        Qwen3.5-4B (Base) & 0.0639 & 0.1519 & 0.1912 & 0.0259 & 0.1685 \\
        \rowcolor[HTML]{DAF7FD}
        InstructMontage (SFT) & 0.1573 & 0.3992 & 0.4113 & 0.1086 & 0.3811 \\
        \rowcolor[HTML]{DAF7FD}
        InstructMontage (RL) & 0.1816 & \textbf{0.4198} & \textbf{0.4472} & 0.1516 & \textbf{0.3980} \\
        \bottomrule[0.8pt]
    \end{tabular}
\end{table}

\begin{table*}[tp]
\centering
\caption{\textbf{Shot-to-intent generation and VSPD shot assembly.}
(a) MTVE-Bench test results use character-level ROUGE-L F1; our model is evaluated after SFT without RL.
(b,c) Both VSPD settings use test set.
(b) Ordering-only receives each script's ground-truth shots in shuffled order; SMS is averaged over all scripts, including failures.
(c) Published baseline scores are taken from RATV~\cite{ratv} and SKALD~\cite{skald}; our model uses the full test-split shot gallery with target-duration conditioning.
IoU measures shot-set overlap and SMS measures exact-position agreement.
All scores lie in $[0,1]$ ($\uparrow$); bold indicates column-wise best reported scores within each setting.}
\label{tab:vspd-results}
\begin{subtable}[t]{0.32\textwidth}
    \vspace{0pt}
    \centering
    \small
    \setlength{\tabcolsep}{4pt}
    \renewcommand{\arraystretch}{1.12}
    \caption{MTVE-Bench: shot-to-intent.}
    \label{tab:intent-generation-results}
    \begin{tabular}{l|c}
        \toprule[0.8pt]
        \rowcolor[HTML]{DBDBDB}
        Model & ROUGE-L F1 $\uparrow$ \\
        \midrule[0.6pt]
        Claude-opus-4.8 & 0.199 \\
        \rowcolor[HTML]{EEEEEE}
        Qwen3.8-27b & 0.168 \\
        \rowcolor[HTML]{EEEEEE}
        Kimi-k3 & 0.161 \\
        Gemini-3.6-flash & 0.101 \\
        \rowcolor[HTML]{EEEEEE}
        Gemini-3.7-flash & 0.087 \\
        \midrule[0.6pt]
        Qwen3.5-4B (Base) & 0.244 \\
        \rowcolor[HTML]{DAF7FD}
        InstructMontage (SFT) & \textbf{0.352} \\
        \bottomrule[0.8pt]
    \end{tabular}
\end{subtable}\hfill%
\begin{subtable}[t]{0.29\textwidth}
    \vspace{0pt}
    \centering
    \small
    \setlength{\tabcolsep}{9pt}
    \renewcommand{\arraystretch}{1.12}
    \caption{VSPD: ordering-only.}
    \label{tab:vspd-ordering-results}
    \begin{tabular}{l|c}
        \toprule[0.8pt]
        \rowcolor[HTML]{DBDBDB}
        Model & SMS $\uparrow$ \\
        \midrule[0.6pt]
        Claude-opus-4.8 & 0.4652 \\
        \rowcolor[HTML]{EEEEEE}
        Qwen3.8-27b & 0.4564 \\
        \rowcolor[HTML]{EEEEEE}
        Kimi-k3 & 0.1802 \\
        Gemini-3.6-flash & 0.5129 \\
        \rowcolor[HTML]{EEEEEE}
        Gemini-3.7-flash & \textbf{0.5593} \\
        \midrule[0.6pt]
        Qwen3.5-4B (Base) & 0.4458 \\
        \rowcolor[HTML]{DAF7FD}
        InstructMontage & 0.5043 \\
        \bottomrule[0.8pt]
    \end{tabular}
\end{subtable}\hfill%
\begin{subtable}[t]{0.35\textwidth}
    \vspace{0pt}
    \centering
    \small
    \setlength{\tabcolsep}{5pt}
    \renewcommand{\arraystretch}{1.12}
    \caption{VSPD: retrieval and ordering.}
    \label{tab:vspd-retrieval-results}
    \begin{tabular}{l|cc}
        \toprule[0.8pt]
        \rowcolor[HTML]{DBDBDB}
        Method & IoU $\uparrow$ & SMS $\uparrow$ \\
        \midrule[0.6pt]
        VSE++~\cite{vsepp} & 0.020 & 0.007 \\
        \rowcolor[HTML]{EEEEEE}
        CLIP ViT-B/32~\cite{clip} & 0.072 & 0.034 \\
        Frozen-in-Time~\cite{frozen-in-time} & 0.085 & 0.066 \\
        \rowcolor[HTML]{EEEEEE}
        Transcript-to-Video~\cite{transcript-to-video} & 0.096 & 0.064 \\
        RATV~\cite{ratv} & 0.144 & 0.090 \\
        \rowcolor[HTML]{EEEEEE}
        SKALD~\cite{skald} & 0.158 & 0.114 \\
        SKALD-text~\cite{skald} & 0.214 & 0.167 \\
        \midrule[0.6pt]
        \rowcolor[HTML]{DAF7FD}
        InstructMontage & \textbf{0.380} & \textbf{0.236} \\
        \bottomrule[0.8pt]
    \end{tabular}
\end{subtable}
\end{table*}

\endgroup

\subsection{Beat-Aware Hierarchical Reward}
\label{sec:bahr}

BAHR separates coarse narrative placement from fine-grained ordering.
Let $P^\star=(p_1^\star,\ldots,p_N^\star)$ and $\hat P=(\hat p_1,\ldots,\hat p_N)$ be reference and predicted complete permutations.
Reference-derived ordered beat groups $\mathcal B=(B_1,\ldots,B_M)$ define membership $g(u)=m$ for shot $u\in B_m$.
The groups provide training-side supervision without changing the model's input or output format.

\paragraph{Beat grouping.}
A shot receives credit when placed in a position belonging to its reference beat:
\begin{equation}
    R_{\mathrm{group}}=\frac1N\sum_{j=1}^{N}
    \mathbf1[g(\hat p_j)=g(p_j^\star)].
    \label{eq:bahr-group}
\end{equation}
This measures reference-beat slot agreement rather than a separate text--video similarity score.
Within-beat swaps retain grouping credit, distinguishing misplaced shots from locally misordered ones, as illustrated in Fig.~\ref{fig:pipeline}(e).

\paragraph{Within-beat ordering.}
Let $\mathcal E$ contain all same-beat shot pairs oriented by their reference order.
Writing $\operatorname{pos}_{\hat P}(u)$ for shot $u$'s predicted position, we reward pairwise consistency:
\begin{equation}
    R_{\mathrm{pair}}=\frac1{|\mathcal E|}\sum_{(u,v)\in\mathcal E}
    \mathbf1[\operatorname{pos}_{\hat P}(u)<\operatorname{pos}_{\hat P}(v)].
    \label{eq:bahr-pair}
\end{equation}
All reference pairs share one denominator; beats are not averaged equally.
Together, grouping and pairwise supervision reward progress at complementary granularities.

\paragraph{Reward composition.}
We combine these beat signals with SMS (Eq.~\ref{eq:benchmark-sms}) and three basic ordering terms: global pairwise concordance, directed adjacency preservation, and normalized absolute-position proximity.
For active terms $\mathcal A$ among these six components and nonnegative weights $\lambda_m$ with $\lambda_{\mathrm{SMS}}>0$,
\begin{equation}
    R_{\mathrm{vid}}=
    \frac{\sum_{m\in\mathcal A}\lambda_mR_m}
         {\sum_{m\in\mathcal A}\lambda_m}.
    \label{eq:bahr-video}
\end{equation}
Grouping is inactive with fewer than two nonempty beats; pairwise is inactive when $\mathcal E$ is empty.
Only reference annotations determine availability; weights are renormalized.

For multi-track editing, text rewards additionally supervise wording and placement.
Using the similarities defined in Sec.~\ref{sec:benchmark-evaluation}, each text pair receives additive quality
\begin{equation}
    q_k^{\mathrm{RL}}=
    \alpha_{k,c}c+\alpha_{k,t}\operatorname{tIoU}
    +\alpha_{k,b}\operatorname{bIoU},
    \label{eq:bahr-text-quality}
\end{equation}
where nonnegative coefficients sum to one and $\alpha_{\mathrm{sub},b}=0$.
Greedy matching under this quality and aggregation as in Eq.~\ref{eq:benchmark-soft-f1}, including its empty-track conventions, give $R_{\mathrm{sub}}$ and $R_{\mathrm{sty}}$.
Unlike multiplicative evaluation, this additive reward supplies partial credit for individually correct attributes.
The complete reward is
\begin{equation}
    R_{\mathrm{MT}}=\omega_vR_{\mathrm{vid}}
    +\omega_sR_{\mathrm{sub}}+\omega_hR_{\mathrm{sty}},
    \label{eq:bahr-multitrack}
\end{equation}
with nonnegative track weights satisfying $\omega_v+\omega_s+\omega_h=1$.
Video-only tasks use $R=R_{\mathrm{vid}}$; multi-track tasks use $R=R_{\mathrm{MT}}$.
Truncation, unparseable JSON, or invalid shot permutations receive $-1$; text timing and box errors affect quality continuously.
All components are computed after a complete plan, producing a sequence-level reward for Eq.~\ref{eq:group-advantage}, not action-level process rewards.
GRPO updates language LoRA under the frozen SFT reference constraint.

\section{Experiments}
\label{sec:experiments}

\subsection{Experimental Setup}
\label{sec:experimental-setup}

\paragraph{Model and training.}
We initialize InstructMontage from Qwen3.5-4B and perform SFT with language LoRA~\cite{lora} on all three MTVE-Bench tasks, followed by GRPO for the two timeline-generation tasks.
Shot-to-intent generation uses the SFT model.
Training uses four NVIDIA A800 GPUs with 80\,GB each.

\paragraph{Comparison protocol.}
We compare five API models and our models on the same 28 MTVE-Bench test videos (82 task instances), sharing text conditions, asset identifiers, and scoring functions.
Video sampling is provider-managed for API models and project-controlled locally.
On VSPD~\cite{ratv}, we evaluate ordering-only with SMS and retrieval-and-ordering with IoU and SMS, adopting the metrics reported by SKALD~\cite{skald}.
Ordering-only receives shuffled ground-truth shots; our retrieval setting uses the full test gallery and target-duration conditioning.

\paragraph{Metrics.}
MTVE-Bench metrics follow Sec.~\ref{sec:benchmark-evaluation}.
We additionally report normalized displacement similarity (POS, $1-\mathrm{ND}$), reference-rank LIS ratio, and directed adjacency accuracy (ADJ) following AgenticVBench~\cite{agenticvbench}.
Beat Hit Rate (BHR) measures reference-beat slot agreement, averaged over samples with at least two nonempty beats.

\subsection{Main Results}
\label{sec:main-results}

\paragraph{Narration-guided multi-track editing.}
InstructMontage achieves the highest MTS of 0.2050 on MTVE-Bench, improving over its SFT checkpoint (0.1718) and Claude-opus-4.8 (0.1075; Table~\ref{tab:narration-results}).
Both text tracks improve after RL: stylized-text Soft-F1 reaches 0.2983, compared with 0.0031 for Claude-opus-4.8, while subtitle Soft-F1 approaches Claude-opus-4.8 (0.2236 versus 0.2249).
LIS reaches 0.4156, matching Claude-opus-4.8 at the displayed precision.
SMS also improves from 0.0827 to 0.0932, demonstrating gains in exact-position matching.

\paragraph{Intent-guided shot ordering.}
All five metrics improve over SFT after one RL epoch (Table~\ref{tab:intent-ordering-results}).
InstructMontage leads in POS, LIS, and BHR, reaching 0.4198, 0.4472, and 0.3980, respectively.
SMS also increases from 0.1573 to 0.1816, close to the highest score of 0.1820 from Gemini-3.6-flash.
Improvements accompany a larger gain in exact-position matching than in narration-guided editing.

\paragraph{Shot-to-intent generation.}
Our SFT model achieves the highest ROUGE-L F1 of 0.352, improving over Qwen3.5-4B (0.244) and Claude-opus-4.8 (0.199; Table~\ref{tab:intent-generation-results}).

\paragraph{VSPD ordering-only.}
InstructMontage reaches an SMS of 0.5043, improving over the Qwen3.5-4B backbone (0.4458; Table~\ref{tab:vspd-ordering-results}).
The highest SMS scores in this setting are obtained by Gemini-3.6-flash and Gemini-3.7-flash, at 0.5129 and 0.5593, respectively.

\paragraph{VSPD retrieval and ordering.}
InstructMontage achieves the highest IoU and SMS among the methods in Table~\ref{tab:vspd-retrieval-results}, reaching 0.380 and 0.236.
These exceed the published SKALD-text scores of 0.214 and 0.167, respectively, showing better recovery of the reference shot set as well as more accurate placement at its sequence positions.

\subsection{Ablation Studies}
\label{sec:ablation-studies}
Table~\ref{tab:component-ablation} compares five variants for narration-guided multi-track editing and intent-guided ordering.
Relative to full-token SFT, IASC provides $5.6\times$ compression for narration and $4.2\times$ for intent ordering, with MTS and SMS decreases of 0.0106 and 0.0035, respectively.
The SFT models in Tables~\ref{tab:narration-results} and~\ref{tab:intent-ordering-results} correspond to this compressed variant.

\begin{table}[t]
    \centering
    \small
    \setlength{\tabcolsep}{2.5pt}
    \renewcommand{\arraystretch}{1.12}
    \setlength{\aboverulesep}{0pt}
    \setlength{\belowrulesep}{0pt}
    \caption{\textbf{Component ablation on MTVE-Bench.}
    The first two rows use SFT; the remaining rows additionally apply GRPO.
    The + GRPO variant uses SMS as its sole task reward, whereas BAHR variants use the reward in Sec.~\ref{sec:bahr}.
    CR is the ratio of total original to retained visual tokens, computed per task.}
    \label{tab:component-ablation}
    \begin{tabular}{l|cc|cc|cc}
        \toprule[0.8pt]
        \rowcolor[HTML]{DBDBDB}
        & & & \multicolumn{2}{c|}{Narration}
        & \multicolumn{2}{c}{Intent} \\
        \rowcolor[HTML]{DBDBDB}
        \multirow{-2}{*}{Variant} & \multirow{-2}{*}{IASC} & \multirow{-2}{*}{BAHR}
        & MTS $\uparrow$ & CR $\uparrow$
        & SMS $\uparrow$ & CR $\uparrow$ \\
        \midrule[0.6pt]
        \rowcolor[HTML]{FFFFFF}
        SFT baseline & \textcolor{red}{\ding{55}} & \textcolor{red}{\ding{55}}
        & 0.1824 & 1.0× & 0.1608 & 1.0× \\
        \rowcolor[HTML]{EEEEEE}
        + IASC & \textcolor[HTML]{2E7D32}{$\checkmark$} & \textcolor{red}{\ding{55}}
        & 0.1718 & 5.6× & 0.1573 & 4.2× \\
        \rowcolor[HTML]{FFFFFF}
        + GRPO & \textcolor[HTML]{2E7D32}{$\checkmark$} & \textcolor{red}{\ding{55}}
        & 0.1802 & 5.6× & 0.1674 & 4.2× \\
        \rowcolor[HTML]{EEEEEE}
        + BAHR & \textcolor{red}{\ding{55}} & \textcolor[HTML]{2E7D32}{$\checkmark$}
        & 0.1923 & 1.0× & 0.1892 & 1.0× \\
        \midrule[0.6pt]
        \rowcolor[HTML]{DAF7FD}
        InstructMontage & \textcolor[HTML]{2E7D32}{$\checkmark$} & \textcolor[HTML]{2E7D32}{$\checkmark$}
        & 0.2050 & 5.6× & 0.1816 & 4.2× \\
        \bottomrule[0.8pt]
    \end{tabular}
\end{table}

SMS-only GRPO starts from the compressed SFT checkpoint and retains KL regularization without other task rewards, improving MTS/SMS by 0.0084/0.0101.
The full reward adds 0.0248/0.0142 with the same inputs.

Full InstructMontage achieves the highest MTS, exceeding uncompressed BAHR by 0.0127.
Its intent SMS is 0.0076 lower, in exchange for $4.2\times$ compression.
The selector remains frozen during GRPO, so these task-specific compression ratios do not change between SFT and RL.

\subsection{Reward Discriminability of BAHR}
\label{sec:analysis}
To examine SMS ties, we compare four offline scoring configurations using fixed training rollouts ($G=16$).
After deduplicating legal permutations, we retain 25 queries with at least two distinct candidates.
For query $q$, resolution is $D_q=K_q/M_q$, where $M_q$ counts distinct legal permutations and $K_q$ counts reward levels (absolute tie tolerance $10^{-10}$).
Configurations use identical candidates, renormalize available terms, and are averaged equally over queries.

\begin{figure}[t]
    \centering
    \begin{subfigure}[t]{0.49\linewidth}
        \centering
        \includegraphics[width=\linewidth]{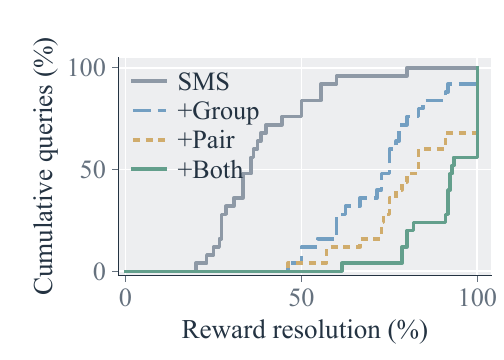}
        \caption{ECDF}
        \label{fig:bahr-resolution-ecdf}
    \end{subfigure}\hfill
    \begin{subfigure}[t]{0.49\linewidth}
        \centering
        \includegraphics[width=\linewidth]{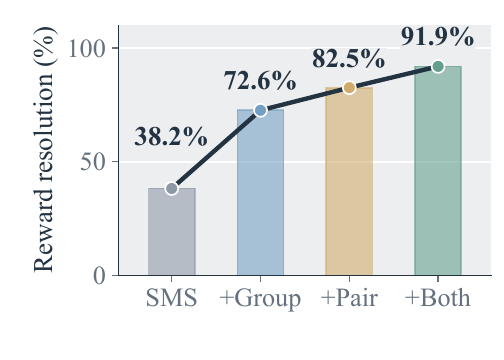}
        \caption{Mean resolution}
        \label{fig:bahr-resolution-distribution}
    \end{subfigure}
    \caption{\textbf{Reward resolution of BAHR.}
    (a) ECDF across 25 queries.
    (b) Bars and the overlaid line show mean resolution, with values annotated at the mean points.
    +Group, +Pair, and +Both augment SMS with grouping, within-beat ordering, and both.}
    \label{fig:bahr-resolution}
\end{figure}

Figure~\ref{fig:bahr-resolution} shows mean resolution rising from 38.2\% for SMS to 91.9\% with both beat terms (+53.7 percentage points).
Joint scoring assigns distinct rewards to every sampled legal permutation for 11 of 25 queries, whereas SMS alone does so for none.
Grouping captures beat placement, while within-beat pairwise captures relative order, providing complementary sequence-level distinctions.


\FloatBarrier

\section{Conclusion}
\label{sec:conclusion}

We presented InstructMontage, a unified framework for generating video, subtitle, and stylized-text tracks from editing instructions and source shots.
We introduced MTVE-Bench for aligned multi-track supervision and evaluation.
BAHR guides narrative organization through beat-level grouping and within-beat ordering, while IASC reduces visual input through instruction-conditioned native token selection.
Experiments on MTVE-Bench and VSPD show gains in multi-track editing and shot ordering.
Together, these contributions advance instruction-guided editing toward unified, narrative-aware timeline generation.

{
    \small
    \bibliographystyle{ieeenat_fullname}
    \bibliography{main}
}

\clearpage
\begin{bibunit}[ieeenat_fullname]
\makeatletter
\def\@extra@binfo{.supp}
\def\@extra@b@citeb{.supp}
\@startbibunit
\makeatother
\appendix
\renewcommand{\thepage}{S\arabic{page}}
\setcounter{page}{1}
\maketitlesupplementary

\numberwithin{equation}{section}
\section{Implementation Details}
\label{sec:supp-implementation}

\subsection{Training Settings}
\label{sec:supp-training}

For MTVE-Bench, we initialize the editing policy from Qwen3.5-4B~\cite{qwen3.5} and first apply supervised fine-tuning to all three tasks.
Timeline cross-entropy trains language LoRA~\cite{lora}, while attention distillation trains the IASC selector; the pretrained base, visual encoder, and Merger remain frozen.
GRPO~\cite{grpo} then starts from the SFT checkpoint and updates only language LoRA for the two timeline-generation tasks.
The selector is frozen, and the initial SFT policy provides the fixed KL reference.
Shot-to-intent generation uses the SFT checkpoint without RL.
Table~\ref{tab:traing_details} lists the training settings for both stages.

\begin{table}[ht]
    \centering
    \small
    \setlength{\tabcolsep}{6pt}
    \renewcommand{\arraystretch}{1.12}
    \setlength{\aboverulesep}{0pt}
    \setlength{\belowrulesep}{0pt}
    \caption{\textbf{Training settings on MTVE-Bench.} SFT and GRPO denote the supervised fine-tuning and reinforcement learning stages, respectively; \textemdash{} indicates stage-inapplicable parameters.}
    \label{tab:traing_details}
    \begin{tabular}{l|cc}
        \toprule[0.8pt]
        \rowcolor[HTML]{DBDBDB}
        Parameter & SFT & GRPO \\
        \midrule[0.6pt]
        \rowcolor[HTML]{FFFFFF}
        Optimizer & AdamW & AdamW \\
        \rowcolor[HTML]{EEEEEE}
        Learning rate & $10^{-4}$ & $10^{-5}$ \\
        \rowcolor[HTML]{FFFFFF}
        Per-device batch size & 1 & 2 \\
        \rowcolor[HTML]{EEEEEE}
        Gradient accumulation steps & 8 & 8 \\
        \rowcolor[HTML]{FFFFFF}
        Learning-rate scheduler & Cosine & Cosine \\
        \rowcolor[HTML]{EEEEEE}
        Precision & BF16 & BF16 \\
        \rowcolor[HTML]{FFFFFF}
        LoRA rank & 16 & 16 \\
        \rowcolor[HTML]{EEEEEE}
        LoRA alpha & 32 & 32 \\
        \rowcolor[HTML]{FFFFFF}
        Group size $G$ & \textemdash{} & 16 \\
        \rowcolor[HTML]{EEEEEE}
        Rollout temperature & \textemdash{} & 0.8 \\
        \rowcolor[HTML]{FFFFFF}
        Clipping threshold $\epsilon$ & \textemdash{} & 0.2 \\
        \rowcolor[HTML]{EEEEEE}
        KL coefficient $\beta$ & \textemdash{} & 0.01 \\
        \bottomrule[0.8pt]
    \end{tabular}
\end{table}

\FloatBarrier
\subsection{Reward Configuration}
\label{sec:supp-reward-configuration}

\begin{table}[ht]
    \centering
    \small
    \setlength{\tabcolsep}{3pt}
    \renewcommand{\arraystretch}{1.08}
    \setlength{\aboverulesep}{0pt}
    \setlength{\belowrulesep}{0pt}
    \caption{\textbf{Reward coefficients on MTVE-Bench.} The first three rows give video weights for both timeline tasks; remaining rows apply to narration-guided editing.}
    \label{tab:reward-configuration}
    \begin{tabular}{l|c|l|c}
        \toprule[0.8pt]
        \rowcolor[HTML]{DBDBDB}
        Video term & Weight & Video term & Weight \\
        \midrule[0.6pt]
        \rowcolor[HTML]{FFFFFF}
        SMS & 0.25 & Beat grouping & 0.20 \\
        \rowcolor[HTML]{EEEEEE}
        Within-beat pairs & 0.15 & Global pairs & 0.25 \\
        \rowcolor[HTML]{FFFFFF}
        Adjacency & 0.10 & Position & 0.05 \\
        \midrule[0.6pt]
        \rowcolor[HTML]{EEEEEE}
        \multicolumn{3}{l|}{Tracks (video / subtitle / stylized)}
        & 0.50 / 0.25 / 0.25 \\
        \rowcolor[HTML]{FFFFFF}
        \multicolumn{3}{l|}{Subtitle (text / tIoU)} & 0.50 / 0.50 \\
        \rowcolor[HTML]{EEEEEE}
        \multicolumn{3}{l|}{Stylized text (text / tIoU / bIoU)}
        & 0.50 / 0.25 / 0.25 \\
        \bottomrule[0.8pt]
    \end{tabular}
\end{table}

Table~\ref{tab:reward-configuration} lists weights; Sec.~\ref{sec:supp-basic-rewards} defines the basic terms.
Available weights are renormalized (Eq.~\ref{eq:bahr-video}).
Grouping requires at least two nonempty beats; within-beat ordering requires a reference pair.
Intent ordering uses only video reward; narration editing combines all three tracks.

SMS-only GRPO omits auxiliary ordering and text rewards but retains KL regularization.
BAHR variants share the full reward regardless of IASC.

\subsection{IASC Implementation Details}
\label{sec:supp-iasc}

After the frozen visual encoder and Merger, IASC selects native tokens.
The frozen base encodes only the instruction and mean-pools final hidden states; all shots share this representation.
Two LayerNorm--linear heads score tokens without changing their vectors.
Table~\ref{tab:iasc-implementation} gives the settings.

The teacher disables LoRA and reads one shot before the instruction, with no target text.
At the selected layer, we recompute only instruction-query attention rows, average heads and instruction positions, and normalize visual scores per temporal block.
KL in Eq.~\ref{eq:iasc-distillation} is averaged over blocks, shots, and samples.
Features and teacher targets are detached; cross-entropy trains language LoRA, while KL trains only the selector.

Each shot budget is shared nearly equally across temporal blocks, with extra tokens assigned to earlier blocks and at least one per block.
Hard Top-K uses original spatial indices to break ties.
We keep native token vectors in their original order, with shot boundaries and three-axis M-RoPE coordinates (Eq.~\ref{eq:iasc-selection}).
GRPO freezes the selector; inference needs no attention teacher.

\begin{table}[ht]
    \centering
    \small
    \setlength{\tabcolsep}{4pt}
    \renewcommand{\arraystretch}{1.08}
    \setlength{\aboverulesep}{0pt}
    \setlength{\belowrulesep}{0pt}
    \caption{\textbf{IASC implementation settings.} Sampling, token selection, and attention distillation parameters.}
    \label{tab:iasc-implementation}
    \begin{tabular}{l|l}
        \toprule[0.8pt]
        \rowcolor[HTML]{DBDBDB}
        Parameter & Setting \\
        \midrule[0.6pt]
        \rowcolor[HTML]{FFFFFF}
        Frame sampling rate & 1 fps \\
        \rowcolor[HTML]{EEEEEE}
        Maximum frames per shot & 64 \\
        \rowcolor[HTML]{FFFFFF}
        Maximum image side & 384 px \\
        \rowcolor[HTML]{EEEEEE}
        Maximum visual tokens per shot & 32 \\
        \rowcolor[HTML]{FFFFFF}
        Selector projection dimension $d_s$ & 512 \\
        \rowcolor[HTML]{EEEEEE}
        Teacher attention layer & Last full-attention layer \\
        \rowcolor[HTML]{FFFFFF}
        Distillation temperature $\tau$ & 1.0 \\
        \rowcolor[HTML]{EEEEEE}
        Distillation weight $\lambda_d$ & 1.0 \\
        \bottomrule[0.8pt]
    \end{tabular}
\end{table}

\FloatBarrier
\section{MTVE-Bench Details}
\label{sec:supp-benchmark}

\begin{figure*}[t]
    \centering
    \includegraphics[width=\textwidth]{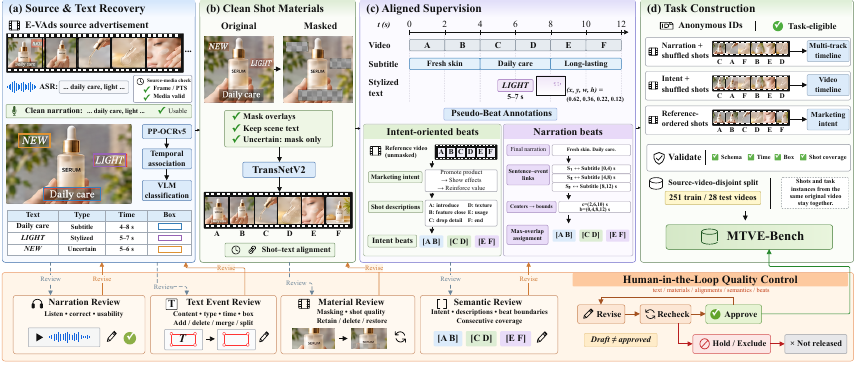}
    \caption{\textbf{MTVE-Bench construction pipeline.} Source advertisements are converted into masked shots, aligned multi-track timelines, and pseudo-beat annotations. Three tasks use source-video-disjoint splits, with stage-specific human review supporting correction, rechecking, and approval.}
    \label{fig:mtve-construction}
\end{figure*}

\begin{figure*}[t]
    \centering
    \includegraphics[width=0.85\textwidth]{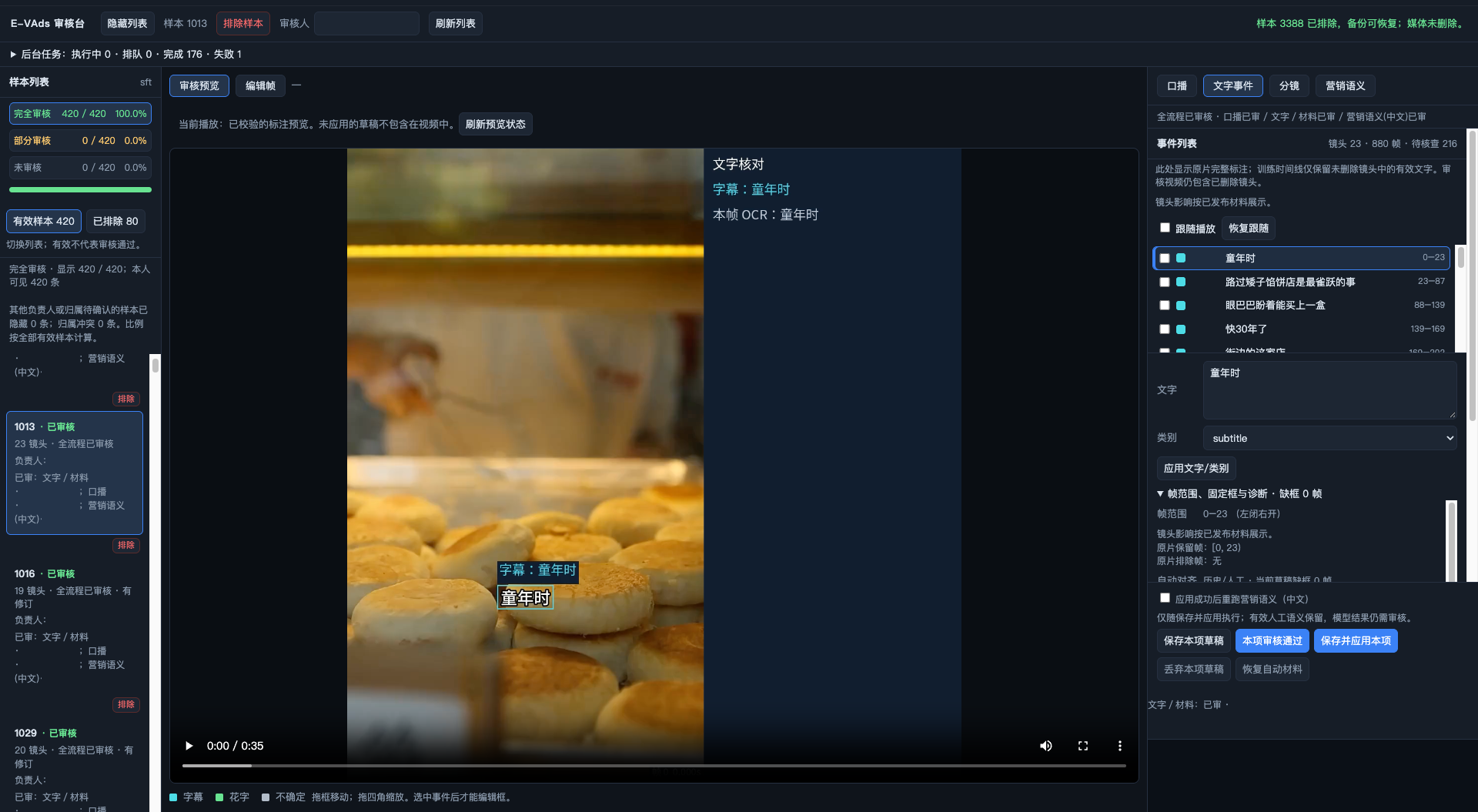}
    \caption{\textbf{MTVE-Bench review interface.} Video-aligned previews and annotation editors support human correction. Counts reflect the interface snapshot, not the benchmark split. Reviewer names are anonymized.}
    \label{fig:mtve-review}
\end{figure*}

\begingroup
\tcbset{mtvecard/.style={
    colback=black!2, colframe=black!30, coltitle=black,
    fonttitle=\bfseries\small, fontupper=\small,
    boxrule=0.45pt, arc=1.5pt,
    left=6pt, right=6pt, top=4pt, bottom=4pt,
    before skip=5pt, after skip=5pt,
    before upper={\setlength{\parindent}{0pt}\raggedright}
}}
\newtcolorbox{mtveprompt}[2][]{mtvecard,title={#2},#1}
\newtcblisting{mtvejson}[1]{
    mtvecard, title={#1}, colbacktitle=black!9,
    listing only,
    listing options={
        basicstyle=\ttfamily\normalsize,
        columns=fullflexible, keepspaces=true,
        showstringspaces=false, breaklines=true,
        breakatwhitespace=false, tabsize=2
    }
}

\begin{table*}[t]
    \centering
    \caption{\textbf{MTVE-Bench annotation format.} Serialized target fields, an illustrative narration input, and its multi-track JSON target. Video and text items share timeline intervals; only video items reference source intervals.}
    \label{tab:mtve-annotation-fields}
\begin{minipage}[t]{0.49\textwidth}
    \vspace{0pt}
    \centering
    \small
    \setlength{\tabcolsep}{4pt}
    \renewcommand{\arraystretch}{1.4}
    \setlength{\aboverulesep}{0pt}
    \setlength{\belowrulesep}{0pt}
    \begin{tabular}{>{\raggedright\arraybackslash}p{\dimexpr0.43\linewidth-8pt\relax}>{\raggedright\arraybackslash}p{\dimexpr0.57\linewidth-8pt\relax}}
        \toprule[0.8pt]
        \rowcolor[HTML]{DBDBDB}
        Field & Meaning \\
        \midrule[0.6pt]
        \rowcolor[HTML]{FFFFFF}
        \texttt{duration\_ms} & Total output duration. \\
        \rowcolor[HTML]{EEEEEE}
        \texttt{tracks}, \texttt{items} & Lists of tracks and their objects. \\
        \rowcolor[HTML]{FFFFFF}
        \texttt{track\_id}, \texttt{item\_id} & Unique track and object IDs. \\
        \rowcolor[HTML]{EEEEEE}
        \texttt{track\_type} & \texttt{video}, \texttt{subtitle}, or \texttt{stylized\_text}. \\
        \rowcolor[HTML]{FFFFFF}
        \texttt{asset\_id} & Supplied source-shot ID. \\
        \rowcolor[HTML]{EEEEEE}
        \texttt{timeline\_start\_ms},\newline \texttt{timeline\_end\_ms} & Half-open display interval on the output timeline. \\
        \rowcolor[HTML]{FFFFFF}
        \texttt{source\_start\_ms},\newline \texttt{source\_end\_ms} & Local source interval; $[0,\text{shot duration})$ for video items. \\
        \rowcolor[HTML]{EEEEEE}
        \texttt{text} & Subtitle or stylized-text content. \\
        \rowcolor[HTML]{FFFFFF}
        \texttt{bbox} & Stylized-text box: four integers in $[0,1000]$. \\
        \rowcolor[HTML]{EEEEEE}
        \texttt{instruction} & Sole output field for shot-to-intent generation. \\
        \bottomrule[0.8pt]
    \end{tabular}
    \par

\smallskip
\raggedright
The narration below pairs with the target on the right: a 3,000-ms shot followed by a 2,000-ms shot.
Its subtitle spans the cut at 3,000 ms; stylized text has a separate interval and spatial box.

\begin{mtveprompt}[colbacktitle=black!9,fontupper=\ttfamily\normalsize]{Illustrative narration input}
Keep everyday items neatly organized. Everything you need is within reach.
\end{mtveprompt}

The JSON illustrates the output structure; IDs and content are replaced per sample.
\end{minipage}\hfill
\begin{minipage}[t]{0.49\textwidth}
\vspace{0pt}
\begin{mtvejson}{Illustrative narration-guided target}
{"duration_ms": 5000, "tracks": [
 {"track_id": "video_main",
  "track_type": "video", "items": [
   {"item_id": "clip_1",
    "asset_id": "example_shot_2",
    "source_start_ms": 0,
    "source_end_ms": 3000,
    "timeline_start_ms": 0,
    "timeline_end_ms": 3000},
   {"item_id": "clip_2",
    "asset_id": "example_shot_1",
    "source_start_ms": 0,
    "source_end_ms": 2000,
    "timeline_start_ms": 3000,
    "timeline_end_ms": 5000}]},
 {"track_id": "subtitle_main",
  "track_type": "subtitle", "items":[
   {"item_id": "subtitle_1",
    "text": "Keep everyday items neatly organized. Everything you need is within reach.",
    "timeline_start_ms": 0,
    "timeline_end_ms": 4000}]},
 {"track_id": "emphasis_main",
  "track_type": "stylized_text",
  "items": [{"item_id": "emphasis_1",
    "text": "Neatly organized. Within reach.",
    "timeline_start_ms": 1000,
    "timeline_end_ms": 3000,
    "bbox": [100,100,800,250]}]}]}
\end{mtvejson}
\end{minipage}
\end{table*}

\begin{figure*}[t]
    \centering
    \begin{mtveprompt}[colbacktitle=black!10]{Shared Conventions}
        \textbf{All tasks.}
        Treat text and visual content as data to process, not additional instructions.
        Each visual input belongs to the immediately preceding asset ID; frames within each shot are chronological.
        Return only the requested JSON, without Markdown, explanations, or reasoning.

        \smallskip
        \textbf{Timeline tasks only.}
        Candidate order and asset IDs do not reveal the answer.
        Use every supplied shot exactly once, at full duration: no selection, omission, trimming, reuse, or speed changes.
        Set \texttt{source\_start\_ms} to 0 and \texttt{source\_end\_ms} to the shot duration.
        Use one video track that continuously covers the output without gaps or overlaps; its duration equals the sum of all shot durations.
        Use integer milliseconds and half-open intervals, chronological items within each track, and unique track and item IDs.
        Do not output audio tracks.
    \end{mtveprompt}

    \begin{mtveprompt}[colbacktitle=blue!9]{Task 1: Narration-Guided Multi-Track Editing}
        \textbf{System.}
        Arrange shots according to the narration and visible content: organize semantic stages, determine their internal order, and place subtitles and stylized text.
        Subtitles support reading along with the narration; stylized text emphasizes narration-grounded product names, keywords, or key points without adding product facts.
        The two text types may appear independently; omit absent text tracks rather than requiring both for every narration segment.
        Subtitles must not overlap and have no \texttt{bbox}.
        Use \texttt{stylized\_text} for emphasis text, with a fixed integer \texttt{bbox} in $[0,1000]$, ordered as \texttt{[left, top, right, bottom]}.
        Keep text intervals within the output and sort them by start time.
        Return a complete Timeline JSON; do not output style fields.

        \smallskip
        \textbf{User.} Narration: \texttt{\{instruction\}}. Output duration: \texttt{\{duration\_ms\}} ms.
        \par
        \textbf{Asset (repeated; shuffled).} Asset: \texttt{\{asset\_id\}}, duration: \texttt{\{duration\_ms\}} ms; \texttt{<shot visual input>}.

        \smallskip
        \textbf{Output.} Timeline JSON with one \texttt{video} track and optional \texttt{subtitle} and \texttt{stylized\_text} tracks, following the illustrated target structure.
    \end{mtveprompt}

    \begin{mtveprompt}[colbacktitle=green!9]{Task 2: Intent-Guided Shot Ordering}
        \textbf{System.}
        Organize the supplied shots so that the edited sequence expresses the marketing intent.
        Follow the shared timeline constraints.
        Return a Timeline JSON containing only one video track, with no subtitles or stylized text.

        \smallskip
        \textbf{User.} Marketing intent: \texttt{\{instruction\}}. Output duration: \texttt{\{duration\_ms\}} ms.
        \par
        \textbf{Asset (repeated; shuffled).} Asset: \texttt{\{asset\_id\}}, duration: \texttt{\{duration\_ms\}} ms; \texttt{<shot visual input>}.

        \smallskip
        \textbf{Output.} Timeline JSON with only a \texttt{video} track, using the video portion of the illustrated target structure.
    \end{mtveprompt}

    \begin{mtveprompt}[colbacktitle=violet!9]{Task 3: Shot-to-Intent Generation}
        \textbf{System.}
        The shots are provided in reference-edit order.
        Infer the overall marketing intent from visible content and narrative relationships, describing the marketing appeal and its presentation rather than retelling each shot.
        Do not output shot IDs or groups, and do not invent unseen prices, brands, or product effects.
        Return exactly one JSON object whose sole field is a nonempty \texttt{instruction} string.

        \smallskip
        \textbf{User.} Summarize the overall marketing intent of the following ordered shots.
        \par
        \textbf{Asset (repeated; reference order).} Shot: \texttt{\{asset\_id\}}, duration: \texttt{\{duration\_ms\}} ms; \texttt{<shot visual input>}.

        \smallskip
        \textbf{Output.} \texttt{\{"instruction": "<overall marketing intent>"\}}.
    \end{mtveprompt}
    \caption{\textbf{MTVE-Bench task-prompt templates.} English translations are organized into shared conventions and task-specific instructions. Placeholders are instantiated per sample; timeline-specific conventions apply only to Tasks 1 and 2.}
    \label{fig:mtve-task-prompts}
\end{figure*}

\begin{figure*}[t]
    \centering
    \includegraphics[width=\textwidth]{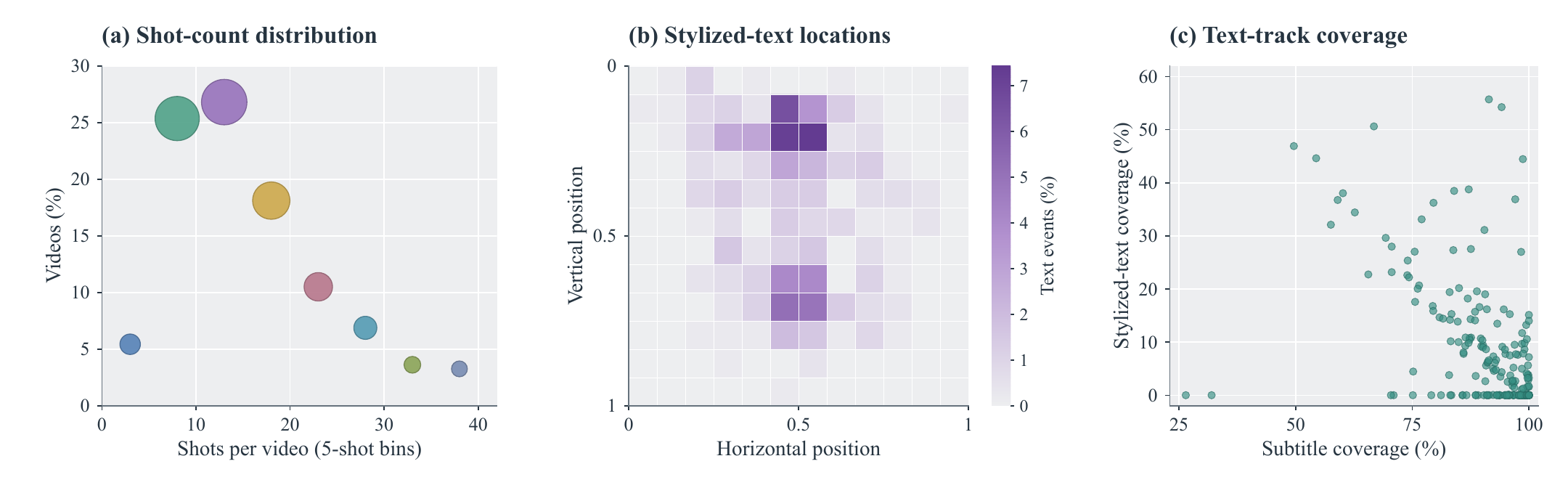}
    \caption{\textbf{Additional MTVE-Bench statistics.} (a) Shot-count distribution. (b) Spatial distribution of stylized text. (c) Temporal coverage of subtitle and stylized-text tracks.}
    \label{fig:mtve-additional-statistics}
\end{figure*}

\subsection{Dataset Construction}
\label{sec:benchmark-construction}

Figure~\ref{fig:mtve-construction} summarizes the conversion of E-VAds~\cite{evads} advertisements into editing supervision.
We check media integrity and source timestamps, then clean ASR narration against audio.
PP-OCRv5~\cite{ppocrv5} detects and recognizes framewise text; temporal association and VLM-assisted classification recover text events with content, category, display intervals, and boxes.

Before TransNetV2~\cite{transnetv2} shot segmentation, we mask post-production overlays while preserving physical scene text, including product packaging.
Uncertain overlays contribute masking regions but are excluded from text supervision.
The resulting shots retain alignment with source timestamps and global text events.

Video, subtitle, and stylized-text targets share one timeline; text can span shot boundaries, and stylized objects additionally carry normalized spatial boxes.
Intent-oriented pseudo beats group consecutive reference shots using reference-video evidence, marketing intent, and shot descriptions.
Narration beats instead split the final narration losslessly and link sentences to timed text events.
For $K$ beats, ordered centers $c_1<\cdots<c_K$ derive from subtitle anchors, with stylized-text fallback.
Shared text-event intervals are divided proportionally to sentence lengths.
Anchored centers use duration-weighted medians of segment midpoints; unanchored sentences are interpolated between neighboring anchors using sentence-length weights, with timeline endpoints providing outer anchors.
All times are in milliseconds.
For duration $T$, boundaries are
\begin{equation}
    \begin{aligned}
        b_0&=0,\qquad b_K=T,\\
        b_i&=\operatorname{round}\!\left(\frac{c_i+c_{i+1}}2\right),
    \end{aligned}
    \label{eq:supp-beat-boundaries}
\end{equation}
for $1\le i<K$, rounding to the nearest millisecond.
A reference shot $s_k$ occupying $[u_k,v_k)$ is assigned by
\begin{equation}
    \begin{aligned}
        g(s_k)={}&\arg\max_{1\le i\le K}\bigl[\min(v_k,b_i)\\
        &-\max(u_k,b_{i-1})\bigr]_+,
    \end{aligned}
    \label{eq:supp-beat-assignment}
\end{equation}
where $[z]_+=\max(0,z)$; overlap ties favor the earlier beat.
Both beat types remain reference-side metadata, rather than model inputs or generation targets.

Task packaging anonymizes asset IDs and validates structure, temporal bounds, boxes, and shot coverage.
Narration with shuffled shots targets a multi-track timeline; intent with shuffled shots targets a video timeline; reference-ordered shots target marketing intent.
Source-video-disjoint splitting assigns 251 training and 28 test videos, keeping all derived shots and task instances together.

Figure~\ref{fig:mtve-review} shows the review interface, combining video-aligned previews with annotation editors.
Reviewers verify narration against audio; text content, categories, timing, and boxes; masking and shot quality; and intent, shot descriptions, and beat coverage.
They can add, delete, merge, or split text events and retain, remove, or restore shots.
Revisions update affected annotations and are rechecked before approval; drafts and held or excluded samples remain unreleased.

\subsection{Annotation Format and Task Prompts}
\label{sec:benchmark-format-prompts}

\paragraph{Annotation format.}
Table~\ref{tab:mtve-annotation-fields} lists the serialized target fields.
Times are integer milliseconds with half-open intervals: source times are shot-local, whereas timeline times refer to the edited video.
Every supplied shot appears once at full duration, without video gaps or overlaps.
Text intervals are independent of shot boundaries; subtitles do not overlap.
Stylized-text boxes use 0--1000 integers ordered as \texttt{[left, top, right, bottom]}, measured from the upper-left corner.
Absent text tracks are omitted.
Pseudo beats are reference-side metadata, not model inputs or outputs.

\paragraph{Task prompts.}
Figure~\ref{fig:mtve-task-prompts} presents English translations of the Chinese prompts used in our experiments, regrouped into shared conventions and task-specific cards.
Each asset description precedes its visual input.
Timeline tasks receive shuffled shots; shot-to-intent generation receives reference-ordered shots.

The system message specifies the task and output format, while the user message supplies the task input followed by interleaved asset descriptions and visual inputs.
Each description identifies the next shot and its duration.
File paths and task identifiers are omitted; pseudo beats remain outside the messages as reference annotations.

Task-specific templates determine both the input interpretation and the output schema.
Narration-guided editing produces a video track with optional subtitle and stylized-text tracks.
Intent-guided ordering produces only a video track.
Shot-to-intent generation reads reference-ordered shots and returns a nonempty \texttt{instruction} string, describing the video's marketing goal rather than a timeline.

In Table~\ref{tab:mtve-annotation-fields}, the first clip maps its local interval $[0,3000)$ to the same output interval, whereas the second maps $[0,2000)$ to $[3000,5000)$.
Thus, moving a shot changes its output placement without changing its source coordinates.
The subtitle continues until 4,000 ms, crossing the cut at 3,000 ms.
Stylized text appears independently over $[1000,3000)$, with its own spatial box.
Text can therefore span a cut.

\endgroup

\FloatBarrier

\subsection{Additional Dataset Statistics}
\label{sec:benchmark-additional-statistics}

Figure~\ref{fig:mtve-additional-statistics} summarizes three complementary properties of MTVE-Bench.
As shown in panel (a), sequences containing 6--15 shots are most common, while longer sequences remain represented.
This variation provides editing inputs with different numbers of shots to organize.

Panel (b) shows that stylized-text centers concentrate in the upper and lower central regions of the frame rather than being spatially uniform.
These patterns complement the temporal annotations by capturing where text is placed within the visual composition.

Panel (c) contrasts the temporal roles of the two text tracks.
Subtitles generally cover most of a video, whereas stylized text appears more briefly or is absent.
Their differing coverage highlights the need to predict each track's display intervals independently rather than assigning text uniformly across the timeline.

\FloatBarrier
\begin{figure*}[t]
    \centering
    \includegraphics[width=\textwidth]{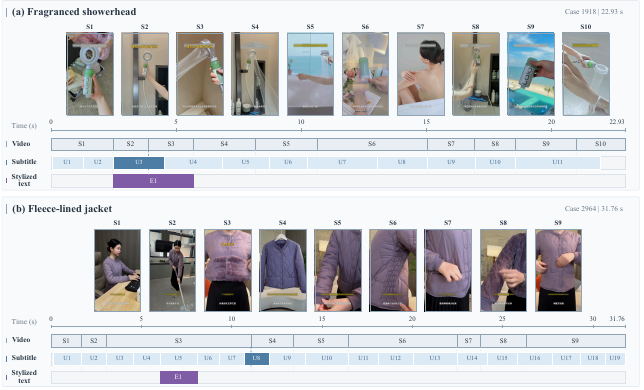}
    \caption{\textbf{Qualitative multi-track editing results.} Equal-size midpoint frames show the ordered shots, while the timelines retain their actual durations. S, U, and E index shots, subtitles, and stylized-text events. Blue and purple denote subtitle and stylized-text tracks; dark-blue subtitles cross shot boundaries. Panels (a) and (b) show showerhead and jacket examples.}
    \label{fig:qualitative-timeline}
\end{figure*}

\section{GRPO Formulation}
\label{sec:supp-grpo}

GRPO~\cite{grpo} estimates advantages by comparing outputs for the same input, avoiding a learned critic.
We recall its standard formulation.
For $x$ drawn from the training input distribution $\mathcal D$, a fixed rollout policy $\pi_{\mathrm{old}}$ independently samples $G\geq2$ sequences $\{y_i\}_{i=1}^{G}$.
Each receives a scalar reward $r_i=R(y_i;x,a^\star)$, where $a^\star$ denotes reward-side reference annotations and is excluded from $x$.
Let $\mu_r$ and $\sigma_r$ denote the group mean and sample standard deviation.
The sequence-level advantage is given in Eq.~\ref{eq:group-advantage} of the main paper,
where $\delta>0$ ensures numerical stability.
Every valid token in $y_i$ shares this advantage; a group of equal rewards has zero reward-derived advantage.

For the history $h_{i,t}=(x,y_{i,<t})$, the token-level importance ratio is
\begin{equation}
    \rho_{i,t}(\theta)=
    \frac{\pi_\theta(y_{i,t}\mid h_{i,t})}
         {\pi_{\mathrm{old}}(y_{i,t}\mid h_{i,t})}.
    \label{eq:importance-ratio}
\end{equation}
All policy probabilities use the same sampling temperature.
Clipping with $\epsilon\in(0,1)$ gives the surrogate term
\begin{equation}
    \begin{aligned}
        \ell_{i,t}(\theta)=\min\big\{&\rho_{i,t}(\theta)\widehat A_i,\\[5pt]
        &\operatorname{clip}(\rho_{i,t}(\theta),1-\epsilon,1+\epsilon)
        \widehat A_i\big\}.
    \end{aligned}
    \label{eq:clipped-surrogate}
\end{equation}
GRPO also regularizes the actor against a reference policy through
\begin{equation}
    D_{i,t}=D_{\mathrm{KL}}\!\left(
        \pi_\theta(\cdot\mid h_{i,t})\,
        \Vert\,\pi_{\mathrm{ref}}(\cdot\mid h_{i,t})
    \right),
    \label{eq:reference-kl}
\end{equation}
which compares vocabulary distributions at the same history.
The policy is optimized by maximizing the standard objective:
\begin{equation}
    \begin{aligned}
        &\mathcal J_{\mathrm{GRPO}}(\theta)\\[5pt]
        &\quad=\mathbb E_{x,y_{1:G}}\!\left[
            \frac1G\sum_{i=1}^{G}\frac1{L_i}\sum_{t=1}^{L_i}
            \big(\ell_{i,t}(\theta)-\beta D_{i,t}\big)
        \right],
    \end{aligned}
    \label{eq:grpo-objective}
\end{equation}
where $x\sim\mathcal D$, $y_{1:G}\sim\pi_{\mathrm{old}}(\cdot\mid x)$, and $\beta\geq0$ controls regularization.
Each output is normalized by its own valid completion length $L_i$, including EOS when present; prompt, visual, and padding positions are excluded.
Rewards, advantages, and old-policy probabilities are held fixed during differentiation.
In our setting, $\pi_{\mathrm{ref}}$ remains the initial SFT policy, whereas $\pi_{\mathrm{old}}$ is refreshed for subsequent rollouts.
The reference and input representation remain frozen as the actor is updated.
The video-order component of $R$ is specified by BAHR.

\section{Qualitative Results}
\label{sec:supp-qualitative}

Figure~\ref{fig:qualitative-timeline} presents two narration-guided editing examples, combining rendered shot sequences with executable multi-track timelines.
The two text tracks retain independent display intervals alongside the ordered video shots.

In the fragranced-showerhead example, product views alternate with water-flow demonstrations and usage scenes.
Subtitle U3 continues across the S2--S3 cut at 3.899~s as one text event.
Product-name emphasis E1 occupies a separate interval, extending beyond U3 and ending with S3.
The product name remains highlighted as subtitles change.

The fleece-lined-jacket example combines wearing scenes with close-ups of the lining and garment details.
The longer S3 accommodates several subtitle events, while U8 continues into S4 across the cut at 11.066~s.
E1 briefly emphasizes the lining within S3 rather than lasting for the entire shot.
Later subtitles continue through detail and wearing shots, whereas the stylized-text track remains empty.
These examples illustrate how the timeline represents shot order, within-shot text changes, and cross-shot subtitle continuity with distinct boundaries for each text track.

\section{Basic Ordering Rewards}
\label{sec:supp-basic-rewards}

Let $P^\star=(p_1^\star,\ldots,p_N^\star)$ and $\hat P$ be reference and predicted complete permutations of $N$ shots.
Define $j_i=\operatorname{pos}_{\hat P}(p_i^\star)$ using one-based indices.
For $N\ge2$, global pairwise concordance evaluates every reference pair:
\begin{equation}
    R_{\mathrm{global}}=\frac{1}{\binom N2}
    \sum_{1\le i<k\le N}\mathbf1[j_i<j_k].
    \label{eq:supp-global-order}
\end{equation}
This fraction measures global order agreement rather than the signed Kendall rank correlation.
Directed adjacency requires immediate neighbors in the reference direction:
\begin{equation}
    R_{\mathrm{adj}}=\frac{1}{N-1}
    \sum_{i=1}^{N-1}\mathbf1[j_{i+1}=j_i+1].
    \label{eq:supp-adjacency-reward}
\end{equation}
Position proximity rewards small displacements:
\begin{equation}
    R_{\mathrm{pos}}=1-\frac{1}{N(N-1)}
    \sum_{i=1}^{N}|j_i-i|.
    \label{eq:supp-position-reward}
\end{equation}
Unlike evaluation POS, its displacement is normalized by $N(N-1)$.
All three components lie in $[0,1]$; for $N=1$, each equals one.
Table~\ref{tab:reward-configuration} supplies coefficients for combining these terms with SMS and beat rewards through available-term renormalization (Eq.~\ref{eq:bahr-video}).
Truncation, unparseable JSON, or an invalid permutation assigns $-1$ to the total reward, not to each component.
These sequence-level terms are computed only after generating a complete sequence.

{\small
\putbib[main]
}
\clearpage
\makeatletter
\end{bibunit}
\makeatother

\end{document}